\documentclass[11pt,a4paper]{article}
\usepackage[hyperref]{naaclhlt2018}
\usepackage{times}
\usepackage{latexsym}
\usepackage{graphicx}
\usepackage{booktabs}
\usepackage{url}
\usepackage{amsmath}
\usepackage{multirow}
\usepackage{adjustbox}
\usepackage{color}
\usepackage{subcaption}
\usepackage{comment}
\usepackage{array}
\newcolumntype{L}[1]{>{\raggedright\arraybackslash}p{#1}}
\usepackage[colorinlistoftodos]{todonotes}
\usepackage{rotating}
\usepackage{fancyhdr}

\DeclareCaptionFont{10pt}{\fontsize{10pt}{12pt}\selectfont}
\aclfinalcopy %
\def\aclpaperid{951} %

\title{Defoiling Foiled Image Captions}

\author{Pranava Madhyastha, Josiah Wang \and Lucia Specia \\
Department of Computer Science \\
University of Sheffield, UK \\
  {\tt {\{p.madhyastha, j.k.wang, l.specia\}@sheffield.ac.uk} }
}

\date{}

\begin{document}
\maketitle

\thispagestyle{fancy}

\begin{abstract}
We address the task of detecting foiled image captions, i.e.\ identifying whether a caption contains a word that has been deliberately replaced by a semantically similar word, thus rendering it inaccurate with respect to the image being described. Solving this problem should in principle require a fine-grained understanding of images to detect linguistically valid perturbations in captions. In such contexts, encoding sufficiently descriptive image information becomes a key challenge. In this paper, we demonstrate that it is possible to solve this task using simple, interpretable yet powerful representations based on explicit object information. Our models achieve state-of-the-art performance on a standard dataset, with scores exceeding those achieved by humans on the task. We also measure the upper-bound performance of our models using gold standard annotations. Our %
analysis reveals that the simpler model performs well even without image information, suggesting that the dataset contains strong linguistic bias.

\end{abstract}

\section{Introduction}

Models tackling vision-to-language (V2L) tasks, for example Image Captioning (IC) and Visual Question Answering (VQA), have demonstrated impressive results in recent years in terms of automatic metric scores. However, whether or not these models are actually learning to address the tasks they are designed for is questionable. 
For example, \newcite{hodosh2016focused} showed that IC models do not understand images sufficiently, as reflected by the generated captions. As a consequence, in the last  few years many diagnostic tasks and datasets have been proposed aiming at investigating the capabilities of such models in more detail to determine whether and how these models are capable of exploiting visual and/or linguistic information ~\cite{shekhar2017foil_acl,johnson2017clevr,antol2015vqa,chen2015deja,gao2015you,yu2015visual,zhu2016visual7w}.

FOIL~\cite{shekhar2017foil_acl} is one such dataset. It was proposed to evaluate the ability of V2L models in understanding the interplay of objects and their attributes in the images and their relations in an image captioning framework. This is done by replacing a word in MSCOCO \cite{lin2014microsoft} captions with a `foiled' word that is semantically similar or related to the original word (substituting \emph{dog} with \emph{cat}), thus rendering the image caption unfaithful to the image content, while yet linguistically valid. 
\newcite{shekhar2017foil_acl} report poor performance for V2L models in classifying captions as foiled (or not). They suggested that %
their models (using image embeddings as input) are very poor at encoding structured visual-linguistic information to spot the mismatch between a foiled caption and the corresponding content depicted in the image.

In this paper, we focus on the foiled captions classification task (Section~\ref{sec:background}), and propose the use of \emph{explicit object detections} as salient image cues for solving the task. In contrast to methods from previous work that make use of word based information extracted from captions~\cite{HeuerSIV2016,yao2015oracle,wu2017image}, we use explicit object category information directly extracted from the images. More specifically, we use an interpretable \emph{bag of objects} as image representation for the classifier. Our hypothesis is that, to truly `understand' the image, V2L models should exploit information about \emph{objects} and their relations in the image and not just global, low-level %
image embeddings as used by most V2L models.

Our main contributions are: 
\begin{enumerate}
\item A model (Section \ref{sec:model}) for foiled captions classification using a simple and interpretable object-based representation, which leads to the best performance in the task (Section \ref{sec:experiments});
\item Insights on upper-bound performance  for foiled captions classification using gold standard object annotations (Section \ref{sec:experiments}); 
\item An analysis of the models, providing insights into the reasons for their strong performance (Section \ref{sec:analysis}).
\end{enumerate}
Our results reveal that the FOIL dataset has a very strong linguistic bias, and that the proposed simple object-based models are capable of finding salient patterns to solve the task. %

\section{Background}
\label{sec:background}

In this section we %
describe the foiled caption classification task and dataset. 

We combine the tasks and data from~\newcite{shekhar2017foil_acl} and~\newcite{shekhar2017b}. 
Given an image and a caption, in both cases the task is to learn a model that can distinguish between a \textbf{REAL} caption that describes the image, and a \textbf{FOIL}ed caption where a word from the original caption is swapped such that it no longer describes the image accurately. There are several sets of `foiled captions' where words from specific parts of speech are swapped: %
\begin{itemize}
\item{\bf{Foiled Noun:}} In this case a noun word in the original caption is replaced with another similar noun, such that the resultant caption is not the correct description for the image. The foiled noun is obtained from list of object annotations from MSCOCO~\cite{lin2014microsoft} and nouns are constrained to the same super-category;
\item {\bf{Foiled Verb:}} Here, verb is foiled with a similar verb. The similar verb is extracted using external resources;
\item {\bf{Foiled Adjective and Adverb:}} Adjectives and adverbs are replaced with similar adjectives and adverbs. Here, the notion of similarity again is obtained from external resources;
\item {\bf{Foiled Preposition:}} Prepositions are directly replaced with functionally similar prepositions.
\end{itemize}

The Verb, Adjective, Adverb and Preposition subsets were obtained using a slightly different methodology (see \newcite{shekhar2017b}) than that used for Nouns \cite{shekhar2017foil_acl}. Therefore, we evaluate these two groups separately. 

\section{Proposed Model}\label{sec:model}

For the foiled caption classification task (Section~\ref{sec:setup}), our proposed model uses information from explicit object detections as an object-based image representation along with textual representations (Section~\ref{sec:representations}) as input to several different classifiers (Section~\ref{sec:classifiers}). %

\subsection{Model definition}
\label{sec:setup}

Let $y \in \{\text{REAL, FOIL}\}$ denote binary class labels. The objective is to learn a model that computes $P(y{\mid}I;C)$, where $I$ and $C$ correspond to the image and caption respectively. %
Our model seeks to maximize a scoring function $\theta$:  
\begin{equation}
y = \arg\max\theta(I;C)
\end{equation}

\subsection{Representations}
\label{sec:representations}

Our scoring function $\theta$ takes in image features and text features (from captions) and concatenates them. We experiment with various types of features. 

For the image side, we propose a \emph{bag of objects} representation for 80 pre-defined MSCOCO categories. We consider two variants:  
(a) \textbf{Object Mention}: A binary vector where we encode the presence/absence of instances of each object category for a given image; 
(b) \textbf{Object Frequency}: A histogram vector where we encode the \emph{number} of instances of each object category in a given image.

For both features, we use \textbf{Gold} MSCOCO object annotations as well as \textbf{Predict}ed object detections using YOLO~\cite{redmon2016yolo9000} pre-trained on MSCOCO to detect instances of the 80 categories.

As comparison, we also compute a standard \textbf{CNN}-based image representation, using the POOL5 layer of a ResNet-152~\cite{he2016deep} CNN pre-trained on ImageNet.
We posit that our object-based representation will better capture semantic information corresponding to the text compared to the CNN embeddings used directly as a feature by most V2L models.

For the language side, we explore two features: 
(a) a simple \textbf{bag of words (BOW)} representation for each caption; (b) an \textbf{LSTM} classifier based model trained on the training part of the dataset. 

Our intuition is that an image description/caption is essentially a result of the interaction between important objects in the image (this includes spatial relations, co-occurrences, etc.).
Thus, representations explicitly encoding object-level information are better suited for the foiled %
caption classification task.

\subsection{Classifiers}
\label{sec:classifiers}

Three types of %
classifiers are explored:
(a) \textbf{Multilayer Perceptron (MLP)}: For BOW-based text representations, a two $100$-dimensional 
hidden layer MLP with ReLU activation function is used with cross-entropy loss, and is optimized with Adam (learning rate 0.001);
(b) \textbf{LSTM Classifier}: For LSTM-based text representations, a uni-directional LSTM classifier is used with 100-dimensional word embeddings and 200-dimensional hidden representations. We train it using cross-entropy loss and optimize it using Adam (learning rate 0.001). Image representations are appended to the final hidden state of the LSTM;
(c) \textbf{Multimodal LSTM (MM-LSTM) Classifier}: As above, except that we initialize the LSTM with the image representation instead of appending it to its output. This can also be seen as am image grounded LSTM based classifier.

\section{Experiments}
\label{sec:experiments}

\begin{table}
\centering
\resizebox{\linewidth}{!}{
\begin{tabular}{*{5}{c}}
\toprule
\textbf{PoS Type} & \textbf{Train$^{foil}$} & \textbf{Test$^{foil}$} & \textbf{Total Train}  & \textbf{Total Test}\\
\midrule
Noun & 153,229 & 75,278  & 306,458 & 150,556\\
Verb & 6,314 & 2,788 & 60,262 & 33,616  \\
Adjective & 15,640 & 9,009 & 73,057  & 42,163 \\  %
Adverb & 1,011 & 451 & 53,381 & 30,738\\  %
Preposition &  8,733& 5,551 & 77,002 & 46,018 \\ %
\bottomrule
\end{tabular}
}
\caption{Dataset statistics for different foiled parts of speech. The superscript ${foil}$ indicates the number of foiled captions.}%
\label{tab:dataset}
\end{table}

\paragraph{Data:}
We use the dataset for nouns from \newcite{shekhar2017foil_acl}\footnote{\url{https://foilunitn.github.io/}} and the datasets for other parts of speech from \newcite{shekhar2017b}
\footnote{The authors have kindly provided us the datasets.}. %
Statistics about the dataset are given in Table~\ref{tab:dataset}.
The evaluation metric is accuracy per class and the average (overall) accuracy over the two classes.

\paragraph{Performance on nouns:}
The results of our experiments with foiled nouns  are summarized in Table~\ref{tab:foilednouns}. First, we note that the models that use \textbf{Gold} bag of objects information are the best performing models across classifiers. 
We also note that the performance is better than human performance. We hypothesize the following reasons for this: (a) human responses were crowd-sourced, which could have resulted in some noisy annotations; (b) our gold object-based features closely resembles the information used for data-generation as described in~\newcite{shekhar2017foil_acl} for the foil noun dataset. %
The models using \textbf{Predict}ed bag of objects from a detector are very close to the performance of \textbf{Gold}.
The performance of models using simple bag of words (\textbf{BOW}) sentence representations and an \textbf{MLP} is better than that of models that use \textbf{LSTM}s. Also, the accuracy of the bag of objects model with \textbf{Frequency} counts is higher than with the binary \textbf{Mention} vector, which only encodes the presence of objects. The Multimodal LSTM (\textbf{MM-LSTM}) has a slightly better performance than \textbf{LSTM} classifiers. In all cases, we observe that the performance is on par with human-level accuracy. Our overall accuracy is substantially higher than that reported in \newcite{shekhar2017foil_acl}. Interestingly, our implementation of \textbf{CNN+LSTM} produced better results than their equivalent model (they reported 61.07\% vs.\ our 87.45\%). We investigate this further in Section~\ref{sec:analysis}.

\begin{table}
\centering
\resizebox{\linewidth}{!}{
\begin{tabular}{*{4}{c}}
\toprule
\textbf{Feats} & \textbf{Overall} & \textbf{Real} & \textbf{Foil}\\
\midrule
Blind (LSTM only)$^{\dag}$ & 55.62 & 86.20 & 25.04 \\
HieCoAtt$^{\dag}$ & 64.14 & 91.89 & 36.38\\
\midrule
CNN + BOW MLP & 88.42 & 86.89 & 89.97 \\
Predict Mention + BOW MLP &  94.94 & 95.68 & 94.23 \\ %
Predict Freq + BOW MLP & 95.14 & 95.82 & 94.48  \\
Gold Mention + BOW MLP & 95.83 & 96.30 & 95.36 \\  %
Gold Freq + BOW MLP & 96.45 & 96.04 & 96.85 \\  %
\midrule
CNN + LSTM & 87.45 & 86.78 & 88.14  \\
Predict Freq + LSTM & 85.99 & 85.17 & 86.81 \\
Gold Freq + LSTM  & 87.38 & 86.62 & 88.18 \\
Predict Freq + MM-LSTM & 87.90 & 86.73 & 88.95 \\
Gold Freq + MM-LSTM & 89.02 & 88.35 & 89.72 \\
\midrule
Human (majority)$^{\dag}$ & 92.89 &  91.24 &  94.52 \\
\bottomrule
\end{tabular}
}
\caption{Accuracy on Nouns dataset. $^{\dag}$ are taken directly from \newcite{shekhar2017foil_acl}. HieCoAtt is the state of the art reported in the paper.}
\label{tab:foilednouns}
\end{table}

\paragraph{Performance on other parts of speech:}
\begin{table}[t]
\centering
\resizebox{\linewidth}{!}{
\begin{tabular}{*{5}{c}}
\toprule
&\textbf{Classifier} & \textbf{Overall} & \textbf{Real} & \textbf{Foil}\\
\midrule
\multirow{3}{*}{\begin{sideways}{VB}\end{sideways}} & Gold Freq + BOW MLP & 84.03 & 97.38 & 70.68 \\  %
& Gold Freq + MM-LSTM & 87.90 & 99.48 & 76.32 \\  %
& HieCoAtt$^{\dag}$ & 81.79 &  - &  57.94 \\
\midrule
\multirow{3}{*}{\begin{sideways}{ADJ}\end{sideways}} & Gold Freq + BOW MLP & 87.74 & 96.96 & 78.52 \\  %
& Gold Freq + MM-LSTM & 92.29 & 85.82 & 98.77 \\  %
& HieCoAtt$^{\dag}$ & 86.00  &  - & 80.05\\
\midrule
\multirow{3}{*}{\begin{sideways}{ADV}\end{sideways}} & Gold Freq + BOW MLP &  54.99 & 98.49 & 11.48 \\ %
& Gold Freq + MM-LSTM & 56.55 & 99.45 & 13.65 \\ %
& HieCoAtt$^{\dag}$ & 53.40  &  -  & 14.73\\
\midrule
\multirow{3}{*}{\begin{sideways}{PREP}\end{sideways}} & Gold Freq + BOW MLP & 75.53 & 92.61 & 58.45  \\ %
& Gold Freq + MM-LSTM & 89.74 & 95.59 & 83.89  \\ %
& HieCoAtt$^{\dag}$ & 74.91  &  - & 61.92\\
\bottomrule
\end{tabular}
}
\caption{Accuracy on Verb, Adjective, Adverb and Preposition datasets, using \textbf{Gold Frequency} as the image representation. $^{\dag}$ is the best performing model as reported in~\newcite{shekhar2017b}.}
\label{tab:posfoiled}
\end{table}

For other parts of speech, we fix the image representation to \textbf{Gold Frequency}, and compare results using the \textbf{BOW}-based \textbf{MLP} and \textbf{MM-LSTM}. We also compare the scores to the state of the art reported in \newcite{shekhar2017b}. Note that this model does not use gold object information and may thus not be directly comparable -- we however recall that only a slight drop in accuracy was found for our models when using predicted object detections rather than gold ones. %
Our findings are summarized in Table~\ref{tab:posfoiled}. The classification performance is not as high as it was for the nouns dataset. Noteworthy is the performance on adverbs, which is significantly lower than the performance across other parts of speech. We hypothesize that this is because of the imbalanced distribution of foiled and real captions in the dataset.
We also found that the performance of \textbf{LSTM}-based models on other parts of speech datasets are almost always better than \textbf{BOW}-based models, indicating the necessity of more sophisticated features.

\section{Analysis}
\label{sec:analysis}

In this section, we attempt to better understand why our models achieve such a high accuracy.

\subsection{Ablation Analysis}
We first perform ablation experiments with our proposed models over the Nouns dataset (FOIL).  We compute image-only models (\textbf{CNN} or \textbf{Gold Frequency}) and text-only models (\textbf{BOW} or \textbf{LSTM}), and investigate which components of our model (text or image/objects) contribute to the strong classification performance (Table~\ref{tab:ablations}). As expected, we cannot classify foiled captions given only image information (global or object-level), resulting in chance-level performance.  

On the other hand, text-only models achieve a very high accuracy. This is a central finding, suggesting that foiled captions are easy to detect even without image information. We also observe that the performance of \textbf{BOW} improves by adding object \textbf{Frequency} image information, but not \textbf{CNN} image embeddings. We posit that this is because there is a tighter correspondence between the bag of objects and bag of word models. In the case of LSTMs, adding either image information helps slightly. The accuracy of our models is substantially higher than that reported in \newcite{shekhar2017foil_acl}, even for equivalent models. %

We note, however, that while the trends of image information is similar for other parts of speech datasets, the performance of \textbf{BOW} based models are lower than the performance of \textbf{LSTM} based models. The anomaly of improved performance of \textbf{BOW} based models seems heavily pronounced in the nouns dataset. %
Thus, we further analyze our model in the next section to shed light on whether the high performance is due to the models or the dataset itself.

\begin{table}[!h]
\centering
\resizebox{0.9\linewidth}{!}{
\begin{tabular}{*{6}{c}}
\toprule
& \textbf{Image} & \textbf{Text} & \textbf{Overall} & \textbf{Real} & \textbf{Foil}\\
\midrule
& CNN & - & 50.01 & 64.71 & 35.31\\
& Gold Freq & - & 50.04 & 53.10 & 47.00 \\
\midrule
\multirow{3}{*}{\begin{sideways}MLP\end{sideways}} 
& - & BOW & 89.33 & 88.32 & 90.34 \\
& CNN & BOW & 88.42 & 86.89 & 89.97\\
& Gold Freq & BOW & 96.45 & 96.04 & 96.85\\
\midrule
\multirow{3}{*}
{\begin{sideways}LSTM\end{sideways}} & - & LSTM & 85.07 & 85.52 & 84.66 \\
& CNN & LSTM & 87.38 & 86.62 & 88.18  \\
& Gold Freq & LSTM & 87.45 & 86.78 & 88.14 \\
\bottomrule
\end{tabular}
}
\caption{Ablation study on FOIL (Nouns).}%
\label{tab:ablations}
\end{table}

\subsection{Feature Importance Analysis}

\label{sec:modelanalysis}
\begin{figure}
\resizebox{\linewidth}{!}{
\includegraphics[width=\linewidth,keepaspectratio]{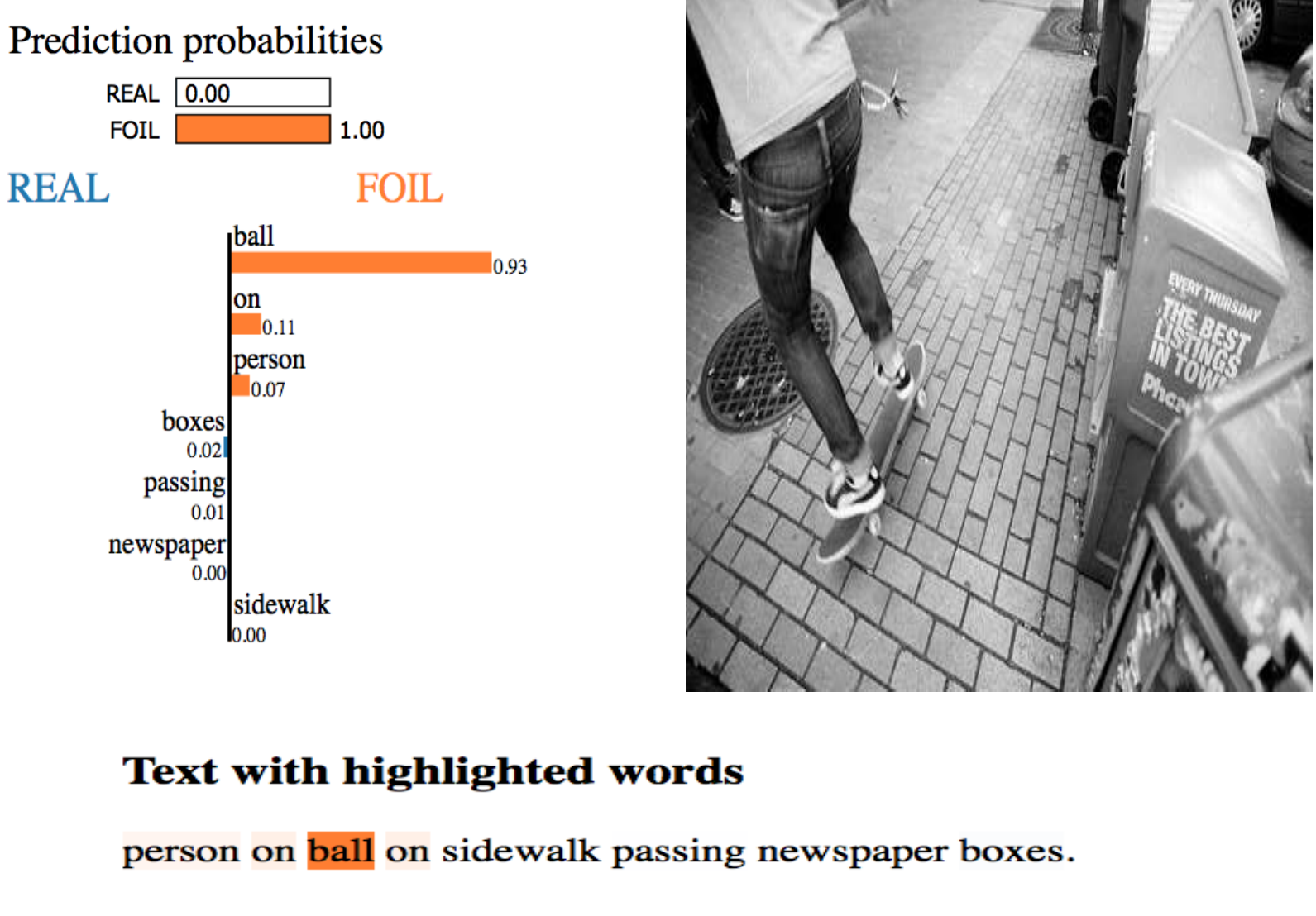}
}
\caption{Classifier's prediction for the foiled caption: The classifier is able to correctly classify the foiled caption and uses the foiled word as the trigger for classification.}
\label{fig:analysis}
\end{figure}

We apply Local Interpretable Model-agnostic Explanations~\cite{ribeiro2016should} to further understand the strong performance of our simple classifier on the Nouns dataset (FOIL) without any image information. We present an example in Figure~\ref{fig:analysis}. We use \textbf{MLP} with \textbf{BOW} only (no image information) as our classifier. As the caption is correctly predicted to be foiled, we observe that the most important feature for classification is the information on the word \emph{ball}, which also happens to be the foiled word. We further analyzed the chances of this happening on the entire test set. We found that 96.56\% of the time the most important classification feature happens to be the foiled word. This firmly indicates that there is a very strong linguistic bias in the training data, despite the claim in~\newcite{shekhar2017foil_acl} that special attention was paid to avoid linguistic biases in the dataset.\footnote{\newcite{shekhar2017foil_acl} have acknowledged about the bias in our personal communications and are currently working on a fix} We note that we were not able to detect the linguistic bias in the other parts of speech datasets.

\section{Conclusions}

We presented an object-based image representation derived from explicit object detectors/gold annotations to tackle the task of classifying foiled captions. The hypothesis was that such models provide the necessary semantic information for the task, while this informaiton is not explicitly present in CNN image embeddings commonly used in V2L tasks. We achieved state-of-the-art performance on the task, and also provided a strong upper-bound using gold annotations. A significant finding is that our simple models, especially for the \emph{foiled noun} dataset, perform well even without  image information. This could be partly due to the strong linguistic bias in the foiled noun dataset, which was revealed by our analysis on our interpretable object-based models. We release our analysis and source code at \url{https://github.com/sheffieldnlp/foildataset.git}.%

\section*{Acknowledgments}
This work is supported by the MultiMT project (H2020 ERC Starting Grant No. 678017). The authors also thank the anonymous reviewers for their valuable feedback on an earlier draft of the paper.

\bibliography{naaclhlt2018_arxiv}
\bibliographystyle{acl_natbib}

\appendix

\end{document}